\documentclass[conference]{IEEEtran}
\IEEEoverridecommandlockouts

\usepackage{graphicx}
\usepackage{caption}
\usepackage[table]{xcolor}
\usepackage[linesnumbered,ruled,vlined]{algorithm2e}
\usepackage{algpseudocode}
\usepackage{algorithmicx}
\usepackage{algcompatible}
\usepackage{amsmath} 
\usepackage{enumitem}
\usepackage{multirow}
\usepackage{subfig}

\definecolor{highlightcolor}{RGB}{255, 255, 180}

\newcommand{\Skip}[1]{}

\newcommand{\ToCheck}[1]{{{#1}}}

\newtheorem{definition}{Definition}
\newtheorem{problem}{Problem}

\begin{document}

\title{\textbf{ECO}: Incremental \textbf{E}go-\textbf{C}entric \textbf{O}ctree Update for Point Streams}

\author{%
  \IEEEauthorblockN{Jaemin Yu$^{\dagger}$, Seongyoon Jeong$^{\dagger}$,
                    Kang-Wook Chon$^{\ast}$, and Duksu Kim$^{\ast}$}
  \IEEEauthorblockA{Korea University of Technology and Education (KOREATECH),
                    Cheonan, Republic of Korea\\[2pt]
    $^{\dagger}$Equal contribution (co-first authors)
    \qquad
    $^{\ast}$Corresponding authors}
  \thanks{This work has been submitted to the IEEE for possible publication.
    Copyright may be transferred without notice, after which this version may
    no longer be accessible.}%
}

\maketitle

\begin{abstract}
Constructing octrees for mobile robots that process continuous point streams in real time poses significant computational and memory challenges.
Standard global structures often suffer from high latency and unbalanced tree growth.
We introduce the Ego-Centric Octree (ECO), a spatial data structure that acts as a 3D sliding window, dynamically bounding the mapping space to the robot's immediate surroundings.
ECO uses an efficient incremental update algorithm that categorizes the environment into shift-out, shift-in, and overlap regions, eliminating redundant global coordinate transformations.
Evaluations on the KITTI benchmark demonstrate that ECO reduces update times by up to \ToCheck{25.60\%} (\ToCheck{24.87\%} on average) compared to full static reconstruction and by up to \ToCheck{67.52\%} (\ToCheck{54.60\%} on average) compared to a bounded incremental baseline.
Furthermore, ECO substantially lowers the total system latency of downstream tasks, running up to \ToCheck{34.17\%} faster than full reconstruction in voxel-map generation.
In dynamic scenes, ECO naturally retains a short-term temporal memory of moving objects, providing useful temporal context while keeping update cost bounded and the tree balanced for real-time spatial perception.
\end{abstract}


\section{Introduction}

An octree is a hierarchical data structure in which each internal node has exactly eight children, used to partition a three-dimensional space by recursively subdividing it into eight octants~\cite{meagher1982octree}.
It serves as an approximate 3D space representation, offering a simple yet efficient method for managing spatial information.
Due to these advantages, octrees are widely employed in various fields that handle 3D data, including simulation, robotics, and autonomous driving~\cite{hou2023,ren2024,miyazaki2025}.
For instance, in mobile robotics, octrees are commonly used to build occupancy grid maps, which are crucial for tasks such as navigation and obstacle avoidance~\cite{asgharivaskasi2023}.

Despite their utility, employing conventional octree construction approaches for mobile robots presents significant challenges.
The primary issue is the high computational overhead associated with building an octree, especially when dealing with large-scale input data such as the dense point clouds generated by LiDAR sensors.
While numerous studies have attempted to accelerate octree construction through methods such as parallel processing and hardware acceleration~\cite{min2021,SimLOD2023}, building an octree on-the-fly for real-time applications such as autonomous driving remains a formidable task~\cite{zhu2024}.
An alternative approach involves pre-processing, where an octree covering the entire target space is built beforehand~\cite{Kai10}. 
However, this method suffers from high spatial overhead, demanding substantial memory resources.
Furthermore, from the perspective of a mobile robot, defining a fixed target space is often impractical.
A robot's operational area can be vast and unpredictable, yet its immediate surroundings are typically the most critical region of interest.
Since the robot's future trajectory is unknown, pre-processing the octree for the path ahead is not a feasible strategy.

To address these limitations, we tackle the problem of efficiently building and maintaining an octree for a mobile robot that perceives its environment as a continuous point stream.
We introduce the \textit{Ego-Centric Octree (ECO)}, a novel concept in which the octree's target space is dynamically bounded to a region of specific dimensions centered precisely on the robot's current position (Sec.~\ref{sec:prob_for}).
This ego-centric approach ensures that computational and memory resources are strictly focused on the most relevant, immediate area.
To complement this spatial structure, we propose an efficient algorithm for incrementally updating the ECO (Sec.~\ref{sec:method}).
By seamlessly incorporating new sensor data and systematically discarding outdated information as the robot navigates, this algorithm dynamically adjusts the octree topology.
This ensures the map remains current, highly localized, and centered within the robot's ego-centric frame without the overhead of global reconstruction.

We evaluated our method on the KITTI benchmark across both static and dynamic environments (Sec.~\ref{sec:result}).
The results demonstrate that the proposed ECO and its incremental update mechanism substantially reduce update cost relative to full reconstruction and global-rooted incremental octrees while keeping the tree shallow and balanced.
Specifically, ECO reduces update times by up to \ToCheck{25.60\%} (\ToCheck{24.87\%} on average) compared to full static reconstructions ($\text{Octree}_{base}$) and by up to \ToCheck{67.52\%} (\ToCheck{54.60\%} on average) compared to a bounded incremental baseline ($\text{i-Octree}_{target}$, a variant of i-Octree~\cite{zhu2024}).
Furthermore, ECO substantially lowers the total system latency of downstream applications, running up to \ToCheck{34.17\%} faster than $\text{Octree}_{base}$ in real-time voxel-map generation.
In dynamic scenarios, ECO operates up to \ToCheck{1.42} times faster than $\text{Octree}_{base}$ while intrinsically preserving a short-term temporal memory of moving objects.
Ultimately, these quantitative gains establish ECO as an efficient, bounded, and balanced octree for real-time local 3D mapping in dynamic robotic applications.

\section{Related Work}

\subsection{Space-Partitioning Data Structures}

Representing and querying 3D spatial data is a fundamental challenge in various applications~\cite{lu2022,hughes2024foundations,10.1145/3329714.3338130}.
To manage this data efficiently, space-partitioning data structures are employed to recursively decompose a space into smaller, non-overlapping regions.
Among the most common are k-d trees, which partition space using axis-aligned hyperplanes~\cite{bentley1975,cai2021ikd}, and Binary Space Partitioning (BSP) trees, which offer more flexibility by using arbitrarily oriented planes~\cite{fuchs1980}.

However, for many real-time applications, the octree has emerged as a particularly effective structure~\cite{yu2021plenoctrees,min2023octomap,zhu2024}.
Its key advantages stem from its regular and hierarchical nature.
This structure enables significant memory compression compared to a uniform voxel grid, as large, uniform areas (either completely empty or fully occupied) can be represented by a single parent node without further subdivision~\cite{museth2013vdb}. 
Concurrently, this hierarchy accelerates spatial queries—such as ray casting for sensor modeling or nearest-neighbor searches for collision detection—by allowing large regions to be processed or dismissed in a single step~\cite{laine2010efficient,rusu20113d,hor2025fast}.
 
These properties of efficiency and memory compression make octrees particularly well-suited for real-time applications on devices with limited computing resources, such as mobile robots and Unmanned Ground Vehicles (UGVs)~\cite{jo2013memory, liang2025real, asgharivaskasi2025riemannian}.

\subsection{Octree Construction}

The octree was first proposed by Meagher as a method for representing 3D objects~\cite{meagher1980octree}, with later work by Gargantini introducing memory-efficient, pointer-less linear variants adapted from quadtrees~\cite{gargantini1982effective}.
The proliferation of modern 3D sensing technologies such as LiDAR has led to massive unstructured point cloud datasets.
This data explosion creates an urgent need for octree construction methods that are both computationally and memory efficient, forming a critical challenge in the field~\cite{1559956}.

To accelerate octree construction, researchers have explored various parallel and out-of-core algorithms~\cite{1559956}.
For instance, to maximize processing speed, Karras et al.~\cite{karras2012maximizing} developed a GPU-based approach that leverages massive parallelism by efficiently sorting Morton-coded point data to rapidly establish the tree's topology.
To address the challenge of massive scale, where datasets exceed system memory, Schütz et al.~\cite{schutz2020fast} proposed a fast out-of-core generation algorithm that streams points from storage, building the tree without loading the entire dataset at once.
Concurrently, to improve memory efficiency, Koh et al.~\cite{koh2020parallel} introduced the truncated octree, a method that processes the point cloud in smaller, parallelized chunks to significantly reduce peak memory usage and achieve higher compression ratios.

While these advanced techniques are powerful, they are fundamentally designed to build a complete octree from a static, pre-captured point cloud representing a fixed volume of space.
In contrast, our work targets the dynamic environment of a mobile platform, such as a UGV equipped with LiDAR.
We focus on an efficient algorithm for incrementally updating an existing octree, which allows the structure to adapt in real-time to the continuous point stream generated by the moving sensor.

\subsection{Incremental Maintenance of Spatial Hierarchies}

In real-time applications such as robotics and autonomous driving, continuously rebuilding spatial data structures such as octrees from scratch is computationally prohibitive.
Incremental update methods, which efficiently modify only affected parts of the structure, are therefore essential for maintaining performance in these dynamic scenes.

This principle has been explored across various data structures~\cite{kopta2012fast, min2023octomap}.
For Bounding Volume Hierarchies (BVHs), common dynamic strategies include node refitting, selective subtree updates based on cost metrics, and asynchronous rebuilding~\cite{ize2007asynchronous, yoon2007ray}.
In the context of k-d trees for robotics, the ikd-tree proposed by Cai et al.~\cite{cai2021ikd} stands out.
It supports incremental updates with newly arriving points and integrates robotics-specific operations such as box-wise manipulation and downsampling.
By avoiding reconstruction, the ikd-tree achieves significant performance gains over static k-d trees.

For octrees, the i-Octree by Zhu et al.~\cite{zhu2024} is an effective dynamic structure for continuously growing maps, significantly reducing construction time by efficiently integrating new points.
While the system is well-suited for persistent, large-scale mapping, its design involves certain trade-offs for mobile robotics applications.
For instance, the continuous growth model can lead to an unbalanced tree structure, affecting long-term query efficiency.
Moreover, for tasks requiring only local awareness, its ever-expanding nature creates unnecessary memory and computational overhead.

Our work differs from prior methods through our ego-centric approach to incremental octree updates.
We maintain a fixed-size, robot-centered map, which naturally bounds memory and computational costs by focusing only on the most relevant space.
Furthermore, our update algorithm is designed to maintain tree balance, ensuring consistently efficient queries.

\section{Ego-Centric Octree and Problem Formulation}
\label{sec:prob_for}

We introduce the \textit{Ego-Centric Octree (ECO)}.
The ECO is an octree representation of 3D space contained within a fixed spatial boundary, defined as a cubic box of side length $L$, whose origin is always anchored to the robot's current position.

This structure has two defining properties.
First, it is \textit{ego-centric}, meaning the map's coordinate system is always relative to the robot and moves with it.
Second, it is \textit{bounded}, as the octree's domain is constrained to a constant, predefined volume.
Conceptually, the ECO acts as a 3D sliding window of the environment that the robot carries, ensuring that computational and memory resources are always focused on the most immediate and relevant space.

\begin{figure*}[t]
    \centering
    \includegraphics[width=\linewidth]{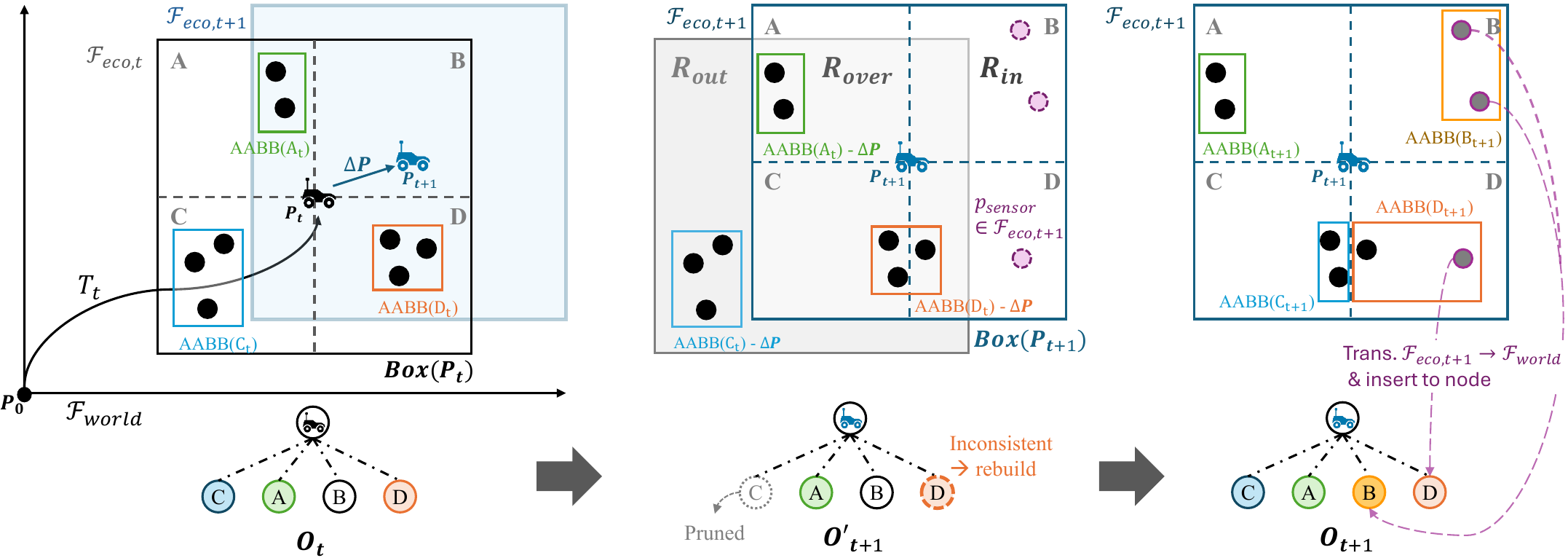}
    \caption{Problem formulation and overview of the incremental ECO update algorithm.
    }
    \label{fig:overview}
\end{figure*}

\Skip{
\begin{figure}[t]
    \centering
    \includegraphics[width=0.8\linewidth]{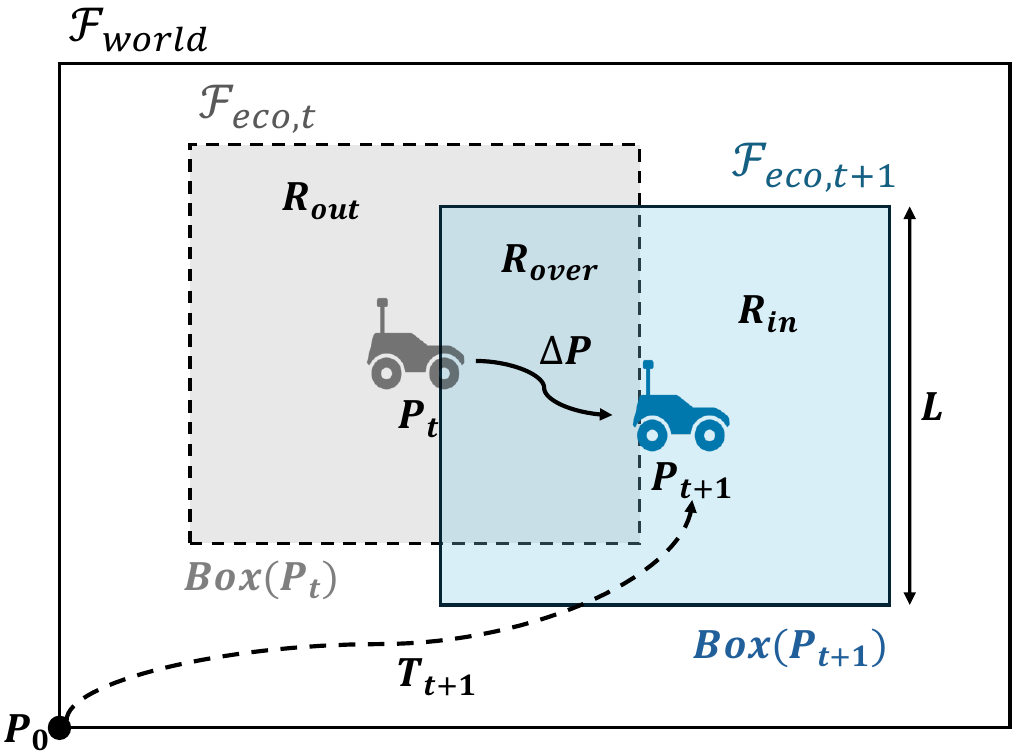}
    \caption{Problem formulation for Ego-Centric Octree (ECO) updates and coordinate systems.
    }
    \label{fig:notatition}
\end{figure}
}

We formulate the problem of maintaining the ECO as a sequential update process.
At each timestep $t$, the system state is defined by the current ECO, denoted as $O_t$, and the robot's corresponding pose (position and orientation), $P_t$.

The inputs required to advance to the next state, $t+1$, are a new \textit{point cloud scan}, $S_{t+1}$, acquired from the robot's sensor, and the robot's new pose, $P_{t+1}$, at which the scan was captured.

The central objective is to compute the subsequent state of the octree, $O_{t+1}$, through an efficient transformation that incorporates this new information. We define this process with the following update function:
\begin{equation}
    O_{t+1} = \text{Update}(O_t, \Delta P, S_{t+1})
    \label{eq:update}
\end{equation}
Here, $\Delta P$ represents the robot's motion (i.e., the transformation from $P_t$ to $P_{t+1}$), which dictates how the existing octree must be shifted before integrating the new points from $S_{t+1}$.

The efficiency of the $\text{Update}$ function stems from deconstructing the operation based on the robot's motion, $\Delta P$. The update is not monolithic; instead, it involves three distinct operations on three spatial volumes defined by the relationship between the ECO's boundaries at times $t$ and $t+1$.

Let $Box(P)$ denote the axis-aligned cubic volume of the ECO with side length $L$, centered at pose $P$. As the robot moves from $P_t$ to $P_{t+1}$, the update process is partitioned across the following three regions (Fig.~\ref{fig:overview}): 
\begin{itemize}
    \item \textbf{Shift-Out Region ($R_{out}$):} The volume $Box(P_t) \setminus Box(P_{t+1})$ that falls outside the new ECO boundary.
    All octree nodes within this region are discarded (\textit{Deletion}).
    \item \textbf{Shift-In Region ($R_{in}$):} The new volume $Box(P_{t+1}) \setminus Box(P_t)$ that has entered the ECO boundary.
    This space is initialized to accommodate new sensor readings (\textit{Initialization}).
    \item \textbf{Overlap Region ($R_{over}$):} The intersecting volume $Box(P_t) \cap Box(P_{t+1})$ where the existing map structure is preserved.
    This region is updated by integrating the new point cloud, $S_{t+1}$, which may reveal previously occluded areas (\textit{Integration}).
\end{itemize}

In the following section, we detail our novel approach to efficiently manage the deletion, initialization, and integration operations across these three distinct regions in real-time while maintaining the balance of the octree, thereby addressing the challenges of ego-centric map maintenance.
\section{Ego-Centric Octree Structure and Incremental Update Algorithm}
\label{sec:method}

In this section, we introduce the ECO structure and its incremental update algorithm.
We begin by detailing the fundamental modifications to a standard octree for our ECO design (Sec.~\ref{subsec:ECO_Structure}).
Following this, we explain the specialized coordinate system and on-demand point update strategy that underpin the ECO's efficiency (Sec.~\ref{subsec:ECO_Coord}).
Finally, we present the comprehensive two-stage algorithm for incrementally updating the ECO with new sensor data (Sec.~\ref{subsec:Incremental_Update}).

\subsection{ECO Structure}
\label{subsec:ECO_Structure}

Fundamentally, the ECO adopts the hierarchical structure of a standard octree, where each internal node has exactly eight children representing a recursive subdivision of its space.
However, to facilitate efficient incremental updates, our ECO structure incorporates two key modifications.

First, each \textit{leaf node} stores an associated set of points that fall within its spatial bounds.
The subdivision process terminates and a node becomes a leaf based on one of two conditions: either the node's volume reaches a predefined minimum resolution, or the number of points it contains falls below a user-defined threshold, $N_{max}$.

Second, every node in the tree—both internal and leaf—stores an \textit{Axis-Aligned Bounding Box (AABB)}.
For a leaf node, this AABB tightly encloses its associated point set.
For an internal node, the AABB is computed in a bottom-up manner by merging the AABBs of its eight children.
This AABB thus represents the complete spatial extent of all points contained within that node's entire subtree, which is crucial for accelerating our update operations.


\subsection{ECO Coordinate System and On-Demand Point Update Strategy}
\label{subsec:ECO_Coord}

To avoid redundant computations and enable our efficient update strategy, the ECO employs a specific coordinate system.
We first define a \textit{world coordinate system}, $\mathcal{F}_{world}$, which is fixed in the world and initialized at the start of operation, with its origin at the robot's initial pose, $P_0$.

The ECO structure itself, however, is dynamically managed relative to the robot's current position.
We define a \textit{local coordinate system} at timestep $t$, $\mathcal{F}_{eco,t}$, whose origin is $P_t$.
The axes of $\mathcal{F}_{eco,t}$ remain parallel to those of $\mathcal{F}_{world}$ as the robot moves, implying a purely translational shift of the local frame relative to the world frame, neglecting rotational effects on the axis-aligned structure.
Within this $\mathcal{F}_{eco,t}$, the ego-volume is defined as a canonical cube, $[-L/2, L/2]$ along each axis.
The transformation matrix, $T_t \in SE(3)$, maps coordinates from $\mathcal{F}_{world}$ to $\mathcal{F}_{eco,t}$ (i.e., $p_{local} = T_t \cdot p_{world}$).
This matrix $T_t$ represents the cumulative ego-motion of the robot, effectively tracking the current pose of the 3D sliding window in $\mathcal{F}_{world}$ (implicitly, as $T_t^{-1}$ would give the pose in the world frame).
As the robot moves from time $t$ to $t+1$ by a relative transformation $\Delta P : \mathcal{F}_{eco,t} \rightarrow \mathcal{F}_{eco,t+1}$, the new transformation matrix is updated as $T_{t+1} = \Delta P \cdot T_t$.

Incoming sensor data ($S_{t+1} \in \mathcal{F}_{eco,t+1}$) is acquired in the robot's current local frame.
All points are transformed into $\mathcal{F}_{world}$ when stored in the ECO; specifically, $p_{stored} = T_{t+1}^{-1} \cdot p_{sensor}$ (where $p_{sensor}$ is in $\mathcal{F}_{eco,t+1}$).
A crucial aspect of our lazy update approach is that while $T_t$ is continuously updated as the robot moves, the coordinates of the points are always stored in $\mathcal{F}_{world}$.
This strategy circumvents the computationally expensive and often redundant task of exhaustively re-transforming all points at every frame.
Instead, points are stored in $\mathcal{F}_{world}$, and for many octree operations—such as traversal to locate specific regions or to determine node boundaries—only the AABBs of the nodes are required.

AABBs of nodes are defined and stored in $\mathcal{F}_{eco,t}$ (the local frame).
These AABBs are implicitly derived from the world-frame points they contain. 
At each timestep, as the robot moves by $\Delta P$, the entire octree structure (including all node AABBs) must dynamically adjust to maintain its definition within the new $\mathcal{F}_{eco,t+1}$.
Since points in $\mathcal{F}_{world}$ remain static, their coordinates relative to the local frame effectively undergo the $\Delta P$ transformation.
Therefore, the stored AABBs (in $\mathcal{F}_{eco,t}$) are efficiently updated by applying the translational component of $\Delta P$ to their coordinates at each timestep.
This ensures the AABBs accurately enclose their constituent points in the current local frame without needing to re-bound from scratch for every node.
For example, if $\Delta P$ is solely a translation $V_{\Delta P}$, then $min_{new} = min_{old} + V_{\Delta P}$ and $max_{new} = max_{old} + V_{\Delta P}$, where $min$ and $max$ denote the minimum and maximum points representing the AABB.

Consequently, during tree traversal, intermediate nodes are processed solely based on their updated AABBs ($\in \mathcal{F}_{eco,t}$).
Only when a query needs the precise local-frame location of points is the current $T_{t+1}$ temporarily applied to the specific $\mathcal{F}_{world}$ point set.
This on-demand transformation significantly reduces the computational burden, as transformations are only performed on a small subset of points within accessed leaf nodes, and only when explicitly necessary.

\subsection{Incremental ECO Update Algorithm}
\label{subsec:Incremental_Update}

Upon the acquisition of new sensor data $S_{t+1}$ at robot pose $P_{t+1}$, the ECO undergoes an incremental update from its state $O_t$ to $O_{t+1}$. 
This process is logically divided into two primary stages: first, refining the existing ECO structure to reflect the robot's movement, and second, integrating the new sensor data into this refined structure (Fig.~\ref{fig:overview}).

The first stage focuses on refining the existing tree ($O_t \rightarrow O'_{t+1}$).
This begins with the \textbf{ego-motion and structure update}.
The cumulative ego-motion matrix $T_{t+1}$ is updated based on the robot's relative transformation $\Delta P$. 
Concurrently, the new ego-volume, $Box(P_{t+1})$, is established as the canonical cube $[-L/2, L/2]$ within the new $\mathcal{F}_{eco,t+1}$ frame. 
To maintain the ECO's definition in the moving local frame, all node AABBs (stored in $\mathcal{F}_{eco,t}$) are efficiently updated by applying the translational component of $\Delta P$ to their coordinates.

Following this, a crucial step involves \textbf{point pruning for the Shift-Out Region} ($R_{out}$).
The algorithm identifies and removes points that have exited the ego-volume. 
A box-box overlap test is performed between each leaf node's updated AABB (now in $\mathcal{F}_{eco,t+1}$) and $Box(P_{t+1})$.
For leaf nodes whose AABB is entirely outside $Box(P_{t+1})$, their entire point set is removed.
For leaf nodes whose AABB partially overlaps $Box(P_{t+1})$, each point ($p_i$) stored in $\mathcal{F}_{world}$ is individually tested via a point-in-box query: $T_{t+1} \cdot p_i \in Box(P_{t+1})$.
Any points falling outside are then deleted from the leaf's point set.

The next part of the refinement stage involves \textbf{selective refinement for the Overlap Region} ($R_{over}$).
This aims to restore the octree's spatial hierarchy consistency where the ego-volume shifted.
Structural inconsistencies are identified when a node's updated AABB no longer fits its parent's octant (both in $\mathcal{F}_{eco,t+1}$).
The mechanism involves propagating AABB refits upwards and locally rebuilding subtrees (by re-inserting their $\mathcal{F}_{world}$ points) where inconsistencies require it, until consistency is restored or a suitable merging point is found.
If a rebuilt subtree contains fewer than $N_{max}$ points, it is merged into a single leaf node.
This localized rebuilding significantly reduces computational overhead.

The second stage involves integrating new sensor data ($O'_{t+1} \rightarrow O_{t+1}$).
First, for \textbf{point identification}, points from $S_{t+1}$ residing within the Shift-In Region ($R_{in}$) are identified.
No point transformations are required for this initial identification, as $S_{t+1}$ is already in $\mathcal{F}_{eco,t+1}$ and $Box(P_{t+1})$ is defined in $\mathcal{F}_{eco,t+1}$.
Next, for \textbf{insertion into the refined ECO}, these identified $R_{in}$ points are inserted into $O'_{t+1}$.
To accelerate this process, a \textit{movement-aware traversal strategy} leveraging $\Delta P$ is employed.
This strategy estimates an optimal starting depth for tree traversal, avoiding redundant top-down searches from the root.
Specifically, given the translational component $V_{\Delta P}$ of $\Delta P$, we determine the start depth ($d_{start}$) such that it represents the deepest level where node side lengths are still greater than or equal to the translation magnitude.
This allows us to begin the search for a point's insertion location from a more localized region of the tree, significantly speeding up the insertion process for frequently updated areas.
The value of $d_{start}$ is calculated as:
\begin{equation}
\small
d_{start} = \min\left(d_{max}, \max\left(0, \left\lfloor \log_2\left(\frac{L}{\|V_{\Delta P}\|}\right) \right\rfloor \right)\right)
\end{equation}
where $d_{max}$ is the maximum possible octree depth, and $\|V_{\Delta P}\|$ is the magnitude of the translational vector of $\Delta P$.
This ensures that traversal begins at a depth where the potential shift is encompassed within one or a few parent nodes.

Upon reaching a target leaf node, each point $p_{sensor}$ (from $S_{t+1}$, which is in $\mathcal{F}_{eco,t+1}$) is transformed into $\mathcal{F}_{world}$ (as described in the previous section) and stored.
If the number of points in a leaf node exceeds a predefined threshold $N_{max}$, the leaf is split into eight children, and its points are redistributed.
This comprehensive two-stage process culminates in the fully updated Ego-Centric Octree, $O_{t+1}$.

%




\section{Results and Analysis}
\label{sec:result}

To validate the efficacy of our proposed method, we conducted a comprehensive evaluation comparing ECO against three alternative octree construction and maintenance strategies.
The experiments were performed on a Linux workstation equipped with an Intel Core i7-9700K CPU and 128GB of RAM.
To ensure a fair comparison focused strictly on algorithmic efficiency, all methods were executed on a single CPU core.

\textbf{Baselines.}
We compared ECO against the following three baselines to evaluate its performance characteristics:

\begin{itemize}
    \item \textbf{$\text{Octree}_{base}$ (Static Baseline)} performs a full reconstruction of the octree from scratch at every timestep, using only the points contained within the current ego-volume $Box(P_t)$.
    It serves as the ground truth for structural accuracy and query performance.

    \item \textbf{$\text{i-Octree}_{global}$ (Global Baseline)} is the standard i-Octree~\cite{zhu2024}, which incrementally expands the map from the initial robot pose $P_0$.
    This represents the traditional dynamic mapping approach where the map size and memory usage continuously increase as the robot explores the environment.
    We utilized the official open-source C++ implementation~\cite{zhu2024}.

    \item \textbf{$\text{i-Octree}_{target}$ (Bounded Baseline)} is a modified version of the official i-Octree implementation.
    This version restricts the map to points within the current $Box(P_t)$ using the same insertion and deletion points as ECO.

\end{itemize}

Both ECO and $\text{Octree}_{base}$ were implemented in C++.
For the i-Octree variants, the official repository code was adapted as described above.

\begin{table}[t]
\caption{Details of the benchmark sequences, including the world scale and the average number of points within the ego-volume.}
\label{tab:benchmark}
\resizebox{\columnwidth}{!}{%
\begin{tabular}{|l|c|c|}
\hline
Benchmark        & World size (m)   & Avg. points / Ego-Vol. \\ \hline \hline
$Seq00_{static}$  &  $667.99~\times~58.70~\times~639.31$    &  1,458,090.01   \\ \hline
$Seq02_{static}$  &  $753.48~\times~94.74~\times~1072.45$   &  855,439.10   \\ \hline
$Seq08_{static}$  &  $543.50~\times~53.37~\times~308.94$    &  1,169,630.33   \\ \hline \hline
$Seq00_{dynamic}$ &  $667.99~\times~58.70~\times~639.31$    &  84,537.49    \\ \hline
$Seq02_{dynamic}$ &  $753.48~\times~94.74~\times~1072.45$   &  87,528.25    \\ \hline
$Seq08_{dynamic}$ &  $543.50~\times~53.37~\times~308.94$    &  81,778.97   \\ \hline
\end{tabular}%
}
\end{table}

\Skip{
\begin{table}[t]
\caption{The details of benchmark sequence, including world scale, ego-voluem size ($L$), leaf capacity ($N_{max}$), and the average number of points within the ego-volume.
}
\label{tab:benchmark}
\resizebox{\columnwidth}{!}{%
\begin{tabular}{|l|c|c|c|}
\hline
Benchmark        & World size (m)   & $L$ (m), $N_{max}$& Avg. points / Ego-Vol. \\ \hline \hline
$Seq0_{static}$     &  667.99 * 58.70 * 639.31    &  20*20*20, 8      &  1,458,090.01   \\ \hline
$Seq2_{static}$     &  753.48 * 94.74 * 1072.45    &  20*20*20, 8 &  855,439.10   \\ \hline
$Seq8_{static}$     &  543.50 * 53.37 * 308.94    &  20*20*20, 8 &  1,169,630.33   \\ \hline \hline
$Seq0_{dynamic}$     &  667.99 * 58.70 * 639.31  & 20*20*20, 8 &  84,537.49    \\ \hline
$Seq2_{dynamic} $ &  753.48 * 94.74 * 1072.45   & 20*20*20, 8 &  87,528.25    \\ \hline
$Seq8_{dynamic}$    & 543.50 * 53.37 * 308.94  & 20*20*20, 8  &  81,778.97   \\ \hline
\end{tabular}%
}
\end{table}
}

\textbf{Benchmarks.}
We utilized the KITTI odometry benchmark
\cite{Geiger2012CVPR} (Sequence 00, 02, 08) to generate test scenarios, categorized into two environment types to rigorously test the update algorithms:

\begin{itemize}
    \item \textbf{Static environment:} To evaluate the algorithmic overhead in an ideal setting without sensor noise or dynamic occlusions, we aggregated point clouds from each sequence into a dense, static global map and downsampled it.
    In this setup, the world-frame coordinates of points remain constant, ensuring that the set of points within an ego-volume is strictly consistent across timesteps.

    \item \textbf{Dynamic environment:} To validate performance in realistic robotic applications, we used the raw sequential LiDAR scans.
    In this scenario, the point cloud within the ego-volume changes dynamically due to moving objects, sensor noise, and the discrete scanning pattern of the LiDAR as the robot moves.

\end{itemize}

Table~\ref{tab:benchmark} summarizes the details of the benchmark sequences, including the scale of the environment and the average number of points maintained within the ego-volume.
Across all benchmark sequences, the ego-volume side length ($L$) and the maximum leaf capacity ($N_{max}$) were uniformly set to $20$ m (a $20 \times 20 \times 20$ m ego-volume) and 8, respectively.

\subsection{Computational Efficiency of Octree Updates}
\label{subsec:comp_eff}

\begin{table*}[t]
\centering
\caption{Average computational time per timestep (ms).
Update, Voxel, and KNN denote the octree update time, voxelization time, and KNN query time, respectively.
A hyphen ('-') indicates that the corresponding baseline was not evaluated in dynamic environments due to the inherent structural limitations of global incremental trees in handling continuous dynamic shifts (detailed in Sec.~\ref{subsec:comp_eff}).}
\label{tab:comprehensive_time}
\begin{tabular}{c|ccc|ccc|ccc|ccc}
\hline
\multirow{2}{*}{\textbf{Bench.}} & \multicolumn{3}{c|}{\textbf{$\text{Octree}_{base}$}} & \multicolumn{3}{c|}{\textbf{$\text{i-Octree}_{global}$}} & \multicolumn{3}{c|}{\textbf{$\text{i-Octree}_{target}$}} & \multicolumn{3}{c}{\textbf{ECO (Ours)}} \\ \cline{2-13} 
 & \textbf{Update} & \textbf{Voxel} & \textbf{KNN} & \textbf{Update} & \textbf{Voxel} & \textbf{KNN} & \textbf{Update} & \textbf{Voxel} & \textbf{KNN} & \textbf{Update} & \textbf{Voxel} & \textbf{KNN} \\ \hline \hline
\multicolumn{13}{c}{(a) Static environment} \\ \hline
Seq00
& 44.04 & 0.11 & 0.022
& 116.70 & 1184.84 & 0.023
& 57.03 & 7.18 & 0.021
& \textbf{33.43} & 0.14 & 0.022
\\
Seq02
& 45.69 & 0.11 & 0.025
& 202.92 & 1870.85 & 0.025
& 104.68 & 7.17 & 0.024
& \textbf{34.00} & 0.14 & 0.026
\\
Seq08
& 44.83 & 0.11 & 0.024
& 144.70 & 1346.87 & 0.024
& 74.64 & 7.19 & 0.023
& \textbf{33.66} & 0.14 & 0.024
\\ \hline
\multicolumn{13}{c}{(b) Dynamic environment} \\ \hline
Seq00 & 4.88 & 0.14 & 0.012 & - & - & - & - & - & - & \textbf{3.43} & 0.17 & 0.008 \\
Seq02 & 5.15 & 0.15 & 0.002 & - & - & - & - & - & - & \textbf{4.47} & 0.20 & 0.002 \\ 
Seq08 & 4.72 & 0.13 & 0.014 & - & - & - & - & - & - & \textbf{3.82} & 0.15 & 0.009 \\ \hline
\end{tabular}%
\end{table*}

\Skip{
\begin{table}[t]
\centering
\caption{Average octree update time per timestep (ms)}
\label{tab:update_time}
\resizebox{\columnwidth}{!}{%
\begin{tabular}{c|c|c|c|c}
\hline
\textbf{Bench.} & \textbf{$\text{Octree}_{base}$} & \textbf{$\text{i-Octree}_{global}$} & \textbf{$\text{i-Octree}_{target}$} & \textbf{ECO (Ours)} \\ \hline \hline
\multicolumn{5}{c}{(a) Static environment} \\ \hline
Seq00 & 45.69 & 202.92 & 104.68 & \textbf{34.00} \\ 
Seq02 & 44.04 & 116.70 & 57.03 & \textbf{33.43} \\ 
Seq03 & 40.60 & 58.75 & 36.96 & \textbf{33.21} \\ \hline
\multicolumn{5}{c}{(b) Dynamic environment} \\ \hline
Seq00 & 4.88 & - & - & \textbf{3.43} \\ 
Seq02 & 5.15 & - & - & \textbf{4.47} \\ 
Seq03 & 4.72 & - & - & \textbf{0.50} \\ \hline
\end{tabular}%
}
\end{table}
} 

\begin{figure}[t]
    \centering
    \subfloat[Sequence 00]{\includegraphics[width=\columnwidth]{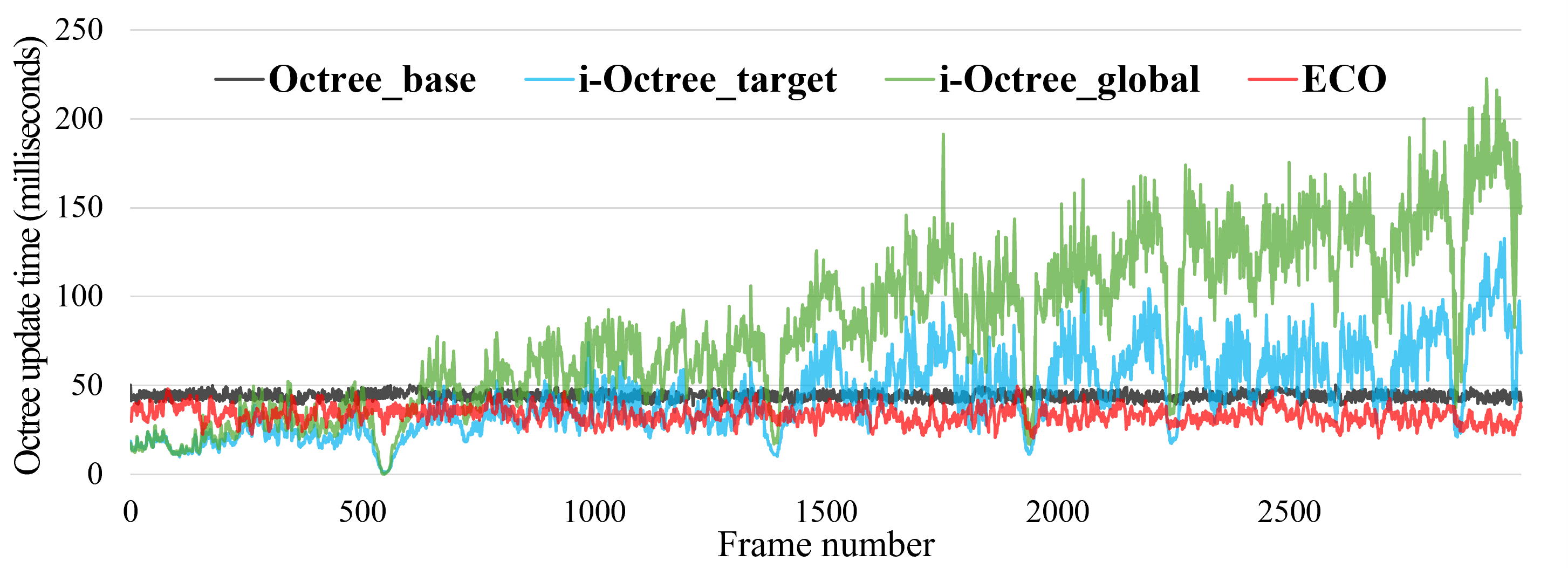}\label{fig:update_time_seq00}}
    \hfill
    \subfloat[Sequence 02]{\includegraphics[width=\columnwidth]{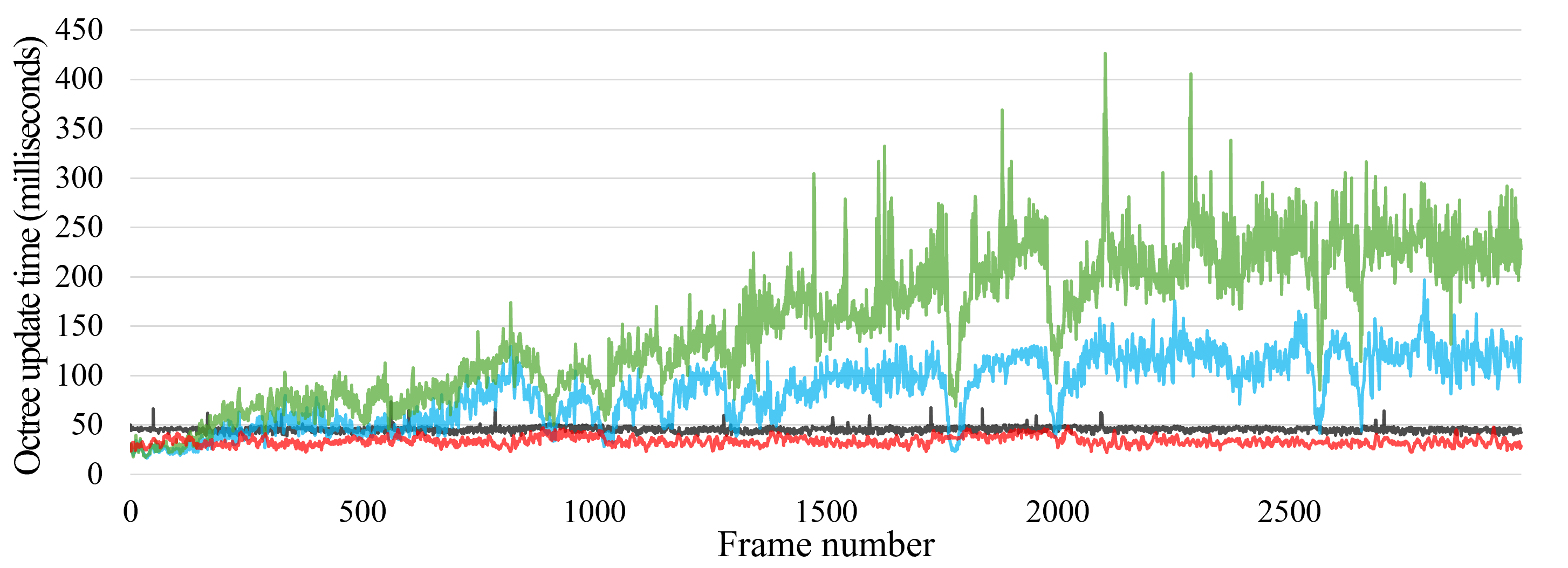}\label{fig:update_time_seq02}}
    \hfill
    \subfloat[Sequence 08]{\includegraphics[width=\columnwidth]{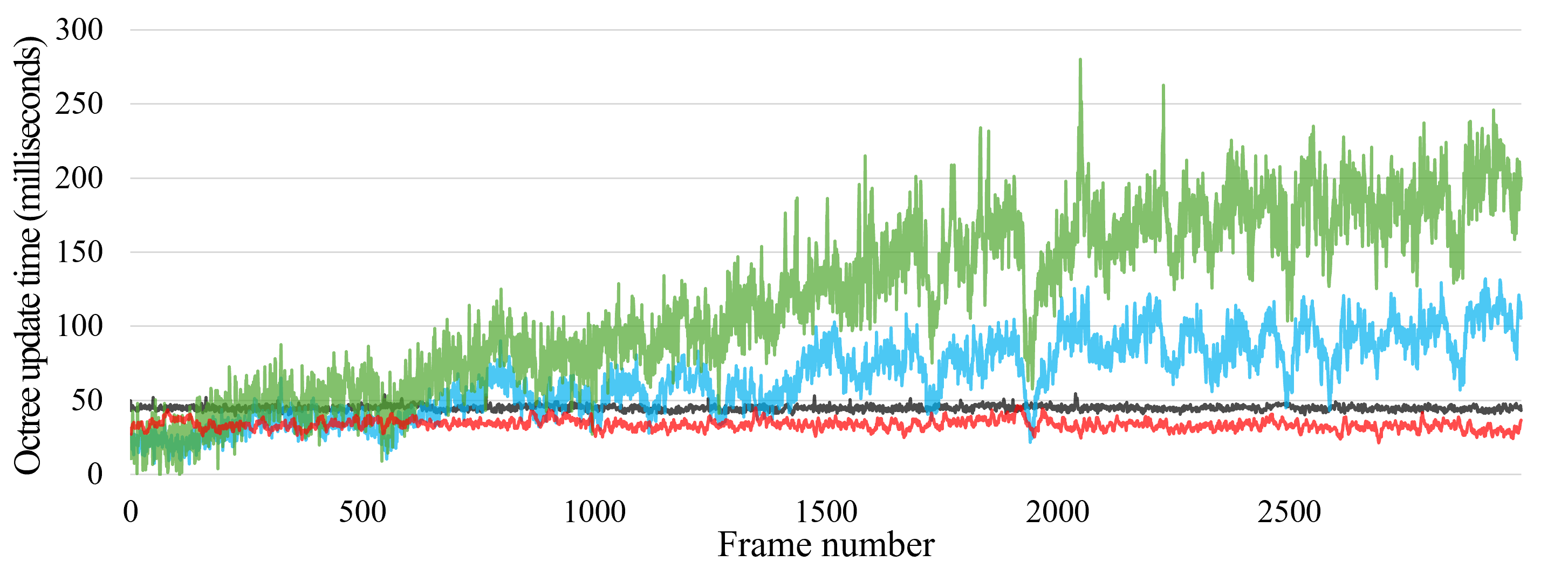}\label{fig:update_time_seq08}}
    \caption{Progression of octree update times over consecutive frames in the static environment benchmarks.
    }
    \label{fig:update_time_static}
\end{figure}

We measured the octree update time for each algorithm, including our proposed method and the baseline alternatives. Table~\ref{tab:comprehensive_time}-(a) summarizes the average update time per timestep, and Fig.~\ref{fig:update_time_static} illustrates the update time progression over the trajectory for the static environment benchmark.

For the standard $\text{i-Octree}_{global}$, the overall octree size (i.e., the total number of points) continuously increases as the robot moves and expands its coverage area. This unbounded growth inherently increases both update time and memory consumption. Consequently, $\text{i-Octree}_{global}$ exhibits significantly lower performance compared to the other methods, taking up to \ToCheck{4.44} times (\ToCheck{3.44} times on average) more time than the full reconstruction baseline, $\text{Octree}_{base}$.

As shown in Fig.~\ref{fig:update_time_static}, $\text{Octree}_{base}$ maintains a highly stable update time across all timesteps. This stability is expected, as the number of points contained within the bounded ego-volume remains relatively consistent throughout the sequence. However, while predictable, this full reconstruction cost is prohibitively high for real-time operation.

On the other hand, the bounded incremental baseline, $\text{i-Octree}_{target}$, initially achieves the shortest update time among all methods. However, as time progresses and the robot traverses further, its update time steadily increases. This performance degradation occurs because the tree's spatial balance is fundamentally broken; the root of the octree remains fixed at the robot's initial position ($P_0$). As a result, updating points at the current robot pose requires significantly more hierarchical traversal steps down to increasingly deeper nodes (e.g., reaching an average depth of \ToCheck{12} levels for $\text{i-Octree}_{target}$, compared to only \ToCheck{5} levels for ECO). This excessive pointer chasing through deep branches severely degrades memory locality and CPU cache efficiency~\cite{museth2013vdb}. This structural inefficiency becomes particularly severe when the robot moves far from the origin, such as in the large-scale trajectory of Sequence 00, heavily penalizing unbalanced incremental structures~\cite{cai2021ikd}.

In contrast, our proposed method, ECO, achieves the smallest average update time among all evaluated methods while simultaneously demonstrating the stable performance characteristic of $\text{Octree}_{base}$. Specifically, ECO requires up to \ToCheck{25.60\%} (\ToCheck{24.87\%} on average) and \ToCheck{67.52\%} (\ToCheck{54.60\%} on average) less update time than $\text{Octree}_{base}$ and $\text{i-Octree}_{target}$, respectively. Although ECO takes slightly longer than the i-Octree methods at the very beginning of the trajectory—due to a slightly more complex initialization and update logic—it quickly outperforms $\text{i-Octree}_{target}$ from a specific crossover point. This sustained efficiency is achieved because our update algorithm actively maintains the tree's spatial balance by strictly defining the current robot position as the center of the root node space, ensuring a bounded, shallow structure that maximizes computational and memory efficiency.

For dynamic environments, utilizing a global-centric structure like the standard i-Octree to maintain a continuously shifting ego-volume is structurally and computationally inefficient. The i-Octree is fundamentally designed for accumulating points into a static global map. Adapting it for dynamic, ego-centric mapping introduces severe preprocessing bottlenecks: it requires continuous, heavy coordinate transformations to align incoming LiDAR frames with the global tree, followed by exhaustive bounding-box filtering to constantly isolate the target local region. Furthermore, extracting and managing a moving volume within a global coordinate system necessitates complex spatial intersection calculations. In contrast, ECO operates entirely within the local frame, efficiently managing updates via overlap regions without redundant global transformations. Most importantly, because the root of a global octree is fixed at the robot's initial position, the tree structure becomes highly unbalanced relative to the current robot pose. ECO directly resolves this by anchoring the tree's root at the current sensor frame, intrinsically guaranteeing shallow, predictable, and bounded query performance.

Due to these inherent limitations of global incremental trees in handling continuous dynamic shifts, our evaluation in dynamic scenes focuses on comparing ECO directly against the full reconstruction baseline, $\text{Octree}_{base}$. Table~\ref{tab:comprehensive_time}-(b) summarizes these results. Consistent with the static environment findings, ECO delivers exceptional performance, running up to \ToCheck{1.42} times (\ToCheck{1.27} times on average) faster than $\text{Octree}_{base}$.

Overall, ECO maintains a stable, bounded per-frame update cost across the trajectory, staying well within real-time requirements even under a limited computational budget.

\subsection{Application Performance Evaluation}

While computational efficiency is critical, a spatial data structure must also demonstrate high utility and accuracy for downstream robotic tasks. We evaluated ECO's performance in two fundamental 3D applications: Voxel Map Construction and K-Nearest Neighbors (KNN) Search.

\subsubsection{\bf Voxel Map Construction}
Generating a localized voxel map is a prerequisite for many navigation and collision-avoidance algorithms. We measured both the computational speed and the structural accuracy (using Chamfer distance against the ground-truth full rebuild) of the generated voxel maps.

Regarding computational speed, the true metric for real-time robotic latency is the \textit{total system time}, defined as the sum of the octree update time and the voxelization query time.

In the static environment benchmark, our results show that querying a perfectly constructed static $\text{Octree}_{base}$ for voxels is extremely fast (\ToCheck{$0.11$} ms on average). While ECO's voxelization query is marginally slower (\ToCheck{$0.14$} ms on average) than the static baseline, it substantially outperforms the bounded global baseline, $\text{i-Octree}_{target}$ (\ToCheck{$7.18$} ms on average). The inefficiency of $\text{i-Octree}_{target}$ stems from the need to traverse a deep, globally anchored tree and perform bounding-box checks at every node to extract the local map. Because ECO is ego-centric, it already perfectly bounds the local space, allowing it to directly map leaf nodes to voxels without wasteful spatial filtering. Consequently, when factoring in the prerequisite octree update times, among the compared octree structures, ECO achieves the lowest total system time for voxel map construction across all static benchmarks, running up to \ToCheck{34.17\%} faster than $\text{Octree}_{base}$.

In the dynamic environment benchmark, the performance trend is similar to the static case. In terms of total system time, ECO consistently outperforms $\text{Octree}_{base}$, operating up to \ToCheck{1.4} times (\ToCheck{1.25} times on average) faster.

Regarding structural accuracy, the voxel map generated by ECO perfectly matches the ground-truth $\text{Octree}_{base}$ in the static benchmark, since the points maintained within the ego-volume are identical.

\begin{figure}[t]
    \centering
    \subfloat[$\text{Octree}_{base}$]{\includegraphics[width=0.3\columnwidth]{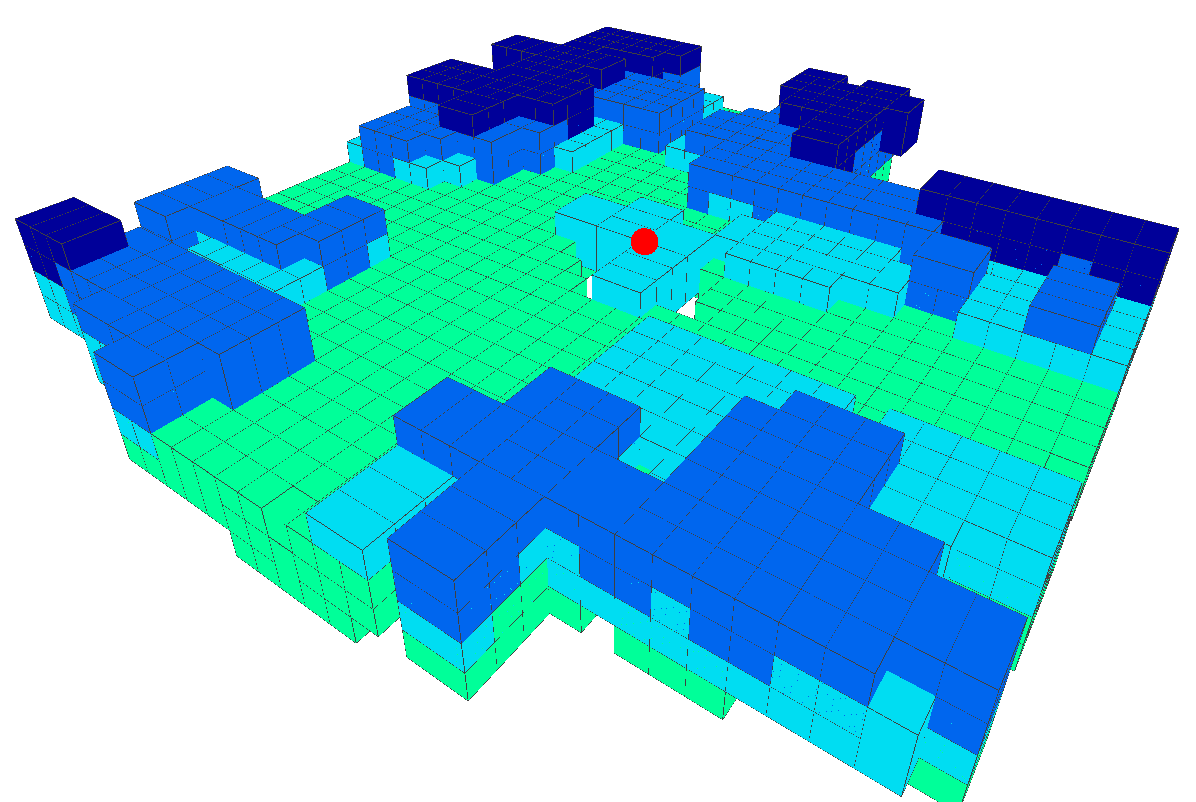}\label{fig:voxel_full}}
    \hfill
    \subfloat[ECO]{\includegraphics[width=0.3\columnwidth]{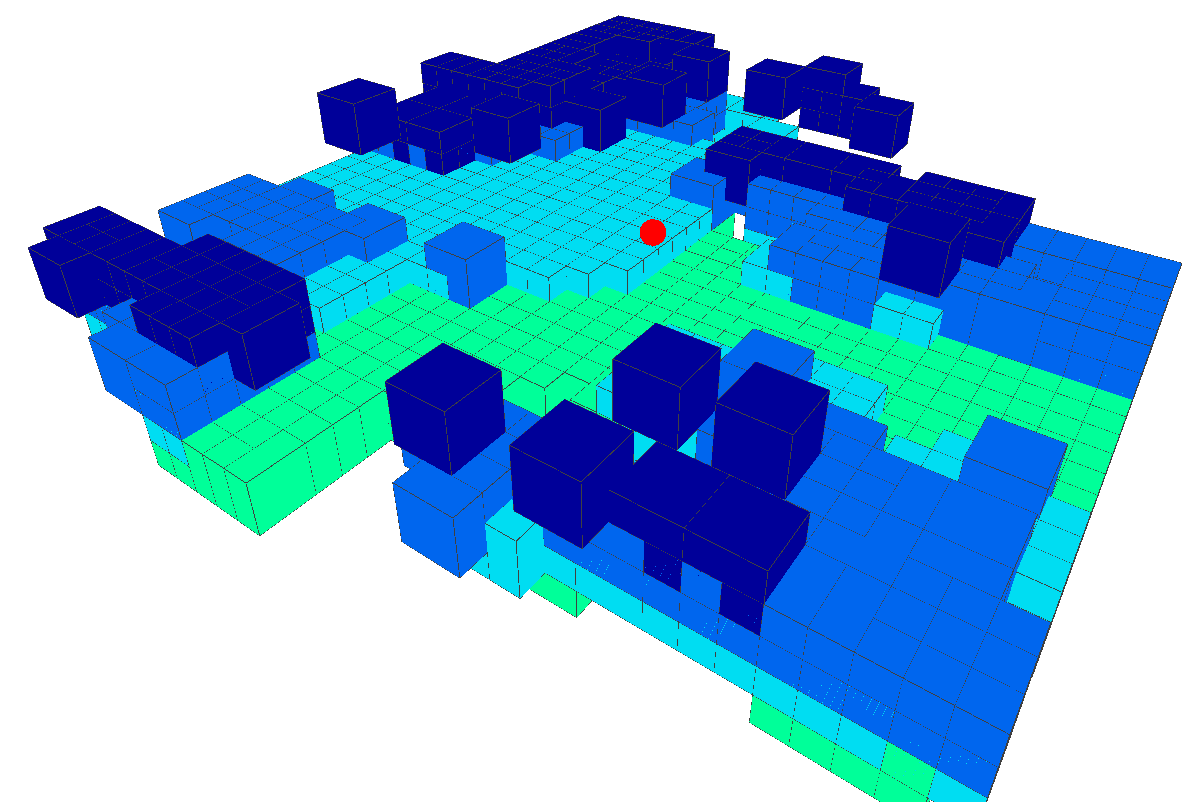}\label{fig:voxel_eco}}
    \hfill
    \subfloat[Error Visualization]{\includegraphics[width=0.32\columnwidth]{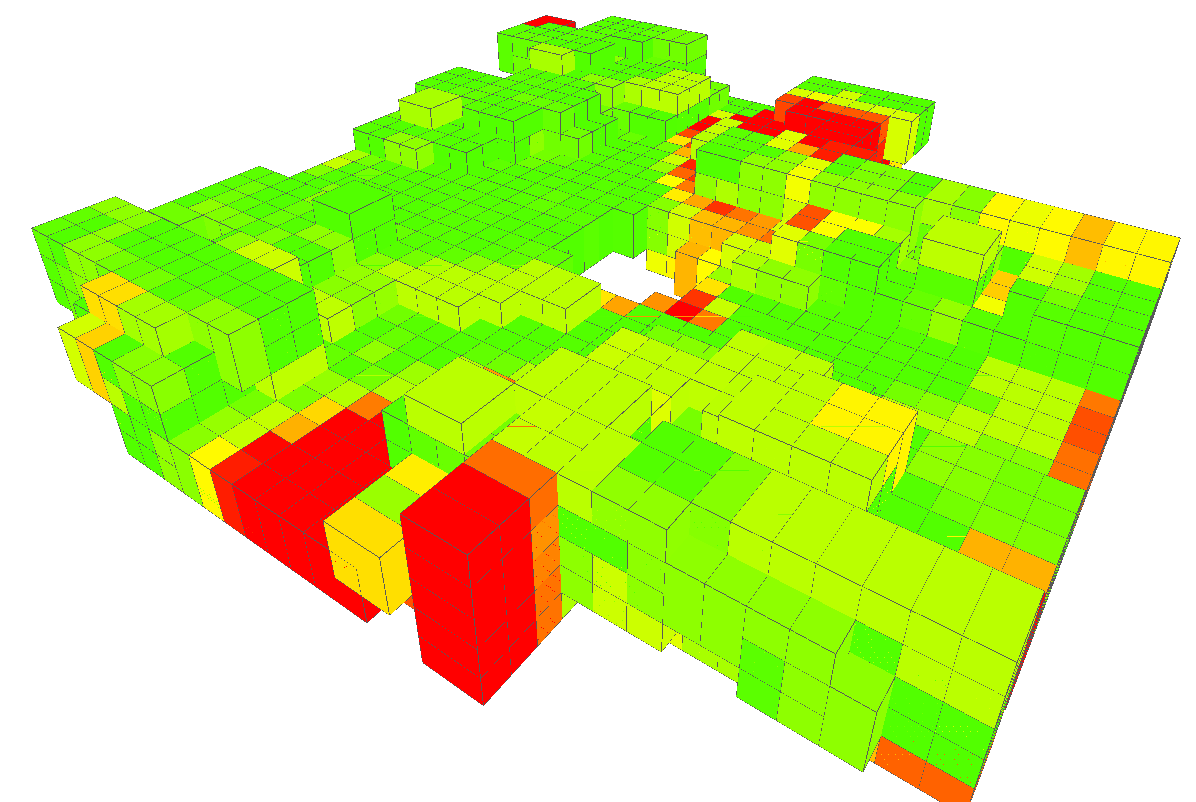}\label{fig:voxel_error}}
    \caption{Qualitative comparison of generated voxel maps in a dynamic environment (frame 104 of Seq00).
    For a continuous real-time comparison, please refer to the supplementary video.
    }
    \label{fig:voxel_comparison}
\end{figure}

In the dynamic environments, we evaluated accuracy by measuring the Chamfer distance between the nearest occupied voxels of the maps generated by $\text{Octree}_{base}$ and ECO (Fig.~\ref{fig:voxel_comparison}).
The Chamfer distance averages approximately \ToCheck{$0.9 \sim 1.2$} m (\ToCheck{$1.07$} m on average across the three sequences). This deviation is not a flaw in the ECO structure, but rather a fundamental characteristic of incremental mapping. While a full rebuild only perceives the instantaneous sensor frame, incremental methods (both ECO and i-Octree) naturally accumulate historical trails of moving objects (e.g., driving cars). Notably, ECO's error profile is identical to that of the standard i-Octree, validating its correctness as an incremental mapping framework.

\subsubsection{\bf KNN Search}
Fast KNN searches are essential for point cloud registration algorithms such as Iterative Closest Point (ICP). Our evaluations using randomized query points demonstrate that the isolated KNN query execution time is remarkably short and comparable across all evaluated methods (\ToCheck{$\sim$0.02} ms in static environments, \ToCheck{$0.002 \sim 0.014$} ms in dynamic scenes).

Since the isolated query time is negligible, the total pipeline latency (update + query) is entirely dominated by the map update phase. Therefore, ECO provides the fastest overall KNN pipeline due to its superior update efficiency. Furthermore, the fact that ECO maintains isolated query times comparable to those of the perfectly balanced $\text{Octree}_{base}$ highlights the core benefit of its ego-centric design. By dynamically re-centering the root node, ECO prevents the tree from becoming skewed or one-sided, preserving an optimal, shallow structure. This ensures bounded $O(\log N)$ search complexity, even after thousands of incremental updates along a trajectory.

In dynamic scenes, the accuracy of the returned neighbors mirrors the traits observed in the voxelization application. Rather than viewing the deviation from the instantaneous full rebuild as an error, this graceful trade-off of single-frame purity for historical accumulation intrinsically provides the map with a short-term temporal memory. By retaining the recent trajectory points of moving objects, ECO provides critical temporal context that is highly valuable for downstream dynamic robotic tasks, such as velocity estimation and predictive obstacle avoidance. Notably, ECO's accuracy profile remains identical to that of the standard i-Octree baseline, validating its correctness as an incremental mapping framework.

\section{Conclusion}
In this paper, we introduced the Ego-Centric Octree (ECO), a spatial data structure that addresses the computational and memory bottlenecks of processing continuous point streams in mobile robotics. By dynamically anchoring a bounded target space to the robot's current position, ECO efficiently focuses resources on the immediate surroundings. To maintain this structure, we proposed an incremental update algorithm that seamlessly manages shift-out, shift-in, and overlap regions as the robot navigates.

Across the evaluated octree-based baselines, ECO provides the lowest total system latency for downstream applications such as voxel mapping and KNN searches, while retaining the exact queries and multi-resolution structure of a hierarchical octree. In dynamic scenes, ECO intrinsically retains a short-term temporal memory by accumulating historical trails of moving objects. This matches the accuracy of standard incremental frameworks while offering valuable temporal context for future tasks, such as velocity estimation and predictive obstacle avoidance. Ultimately, ECO delivers a practical, bounded real-time solution for local spatial awareness.

\textbf{Limitations and Future Work:}
Currently, the ECO structure relies on a fixed cubic boundary and primarily accounts for translational shifts, neglecting rotational effects on the axis-aligned tree. Future work will expand the update algorithm to handle full rotational dynamics within the ego-centric frame without sacrificing efficiency. Additionally, we plan to explore adaptive boundary sizing, enabling the ego-volume to scale dynamically based on the robot's speed and environmental complexity.

\bibliographystyle{IEEEtran}
\bibliography{references}

\end{document}